# Integration of pre-trained protein language models into geometric deep learning networks

Fang Wu [1], Lirong Wu[1], Dragomir Radev[2], Jinbo Xu [3,4] & Stan Z. Li[1✉]

Geometric deep learning has recently achieved great success in non-Euclidean domains, and learning on 3D structures of large biomolecules is emerging as a distinct research area. However, its efficacy is largely constrained due to the limited quantity of structural data. Meanwhile, protein language models trained on substantial 1D sequences have shown burgeoning capabilities with scale in a broad range of applications. Several preceding studies consider combining these different protein modalities to promote the representation power of geometric neural networks but fail to present a comprehensive understanding of their benefits. In this work, we integrate the knowledge learned by well-trained protein language models into several state-of-the-art geometric networks and evaluate a variety of protein representation learning benchmarks, including protein-protein interface prediction, model quality assessment, protein-protein rigid-body docking, and binding affinity prediction. Our findings show an overall improvement of 20% over baselines. Strong evidence indicates that the incorporation of protein language models' knowledge enhances geometric networks' capacity by a significant margin and can be generalized to complex tasks.

[1] AI Research and Innovation Laboratory, Westlake University, 310030 Hangzhou, China. [2] Department of Computer Science, Yale University, New Haven, CT 06511, USA. [3] Institute of AI Industry Research, Tsinghua University, Haidian Street, 100084 Beijing, China. [4] Toyota Technological Institute at Chicago, Chicago, IL 60637, USA. ✉email: stan.zq.li@westlake.edu.cn

Macromolecules (e.g., proteins, RNAs, or DNAs) are essential to biophysical processes. While they can be represented using lower-dimensional representations such as linear sequences (1D) or chemical bond graphs (2D), a more intrinsic and informative form is the three-dimensional geometry[1]. 3D shapes are critical to not only understanding the physical mechanisms of action but also answering a number of questions associated with drug discovery and molecular design[2]. Consequently, tremendous efforts in structural biology have been devoted to deriving insights from their conformations[3–5].

With the rapid advances of deep learning (DL) techniques, it has been an attractive challenge to represent and reason about macromolecules' structures in the 3D space. In particular, different sorts of 3D information, including bond lengths and dihedral angles, play an essential role. In order to encode them, a number of 3D geometric graph neural networks (GGNNs) or CNNs[6–9] have been proposed, and simultaneously achieve several crucial properties of Euclidean geometry such as E(3) or SE(3) equivariance and symmetry. Notably, they are essential constituents of geometric deep learning (GDL), an umbrella term that generalizes networks to Euclidean or non-Euclidean domains[10].

Meanwhile, the anticipated growth of sequencing promises unprecedented data on natural sequence diversity. The abundance of 1D amino acid sequences has spurred increasing interest in developing protein language models at the scale of evolution, such as the series of ESM[11–13] and ProtTrans[14]. These protein language models can capture information about secondary and tertiary structures and can be generalized across a broad range of downstream applications. To be explicit, they have recently been demonstrated with strong capabilities in uncovering protein structures[12], predicting the effect of sequence variation on function[11], learning inverse folding[15] and many other general purposes[13].

With the fruitful progress in protein language models, more and more studies have considered enhancing GGNNs' ability by leveraging the knowledge of those protein language models[12,16,17]. This is nontrivial because compared to sequence learning, 3D structures are much harder to obtain and thus less prevalent. Consequently, learning about the structure of proteins leads to a reduced amount of training data. For example, the SAbDab database[18] merely has 3K antibody-antigen structures without duplicate. The SCOPe database[19] has 226K annotated structures, and the SIFTS database[20] comprises around 220K annotated enzyme structures. These numbers are orders of magnitude lower than the data set sizes that can inspire major breakthroughs in the deep learning community. In contrast, while the Protein Data Bank (PDB)[21] possesses approximately 182K macromolecule structures, databases like Pfam[22] and UniParc[23] contains more than 47M and 250M protein sequences respectively.

In addition to the data size, the benefit of protein sequence to structure learning also has solid evidence and theoretical support. Remarkably, the idea that biological function and structures are documented in the statistics of protein sequences selected through evolution has a long history[24]. The unobserved variables that decide a protein's fitness, including structure, function, and stability, leave a record in the distribution of observed natural sequences[25]. Those protein language models use self-supervision to unlock the information encoded in protein sequence variations, which is also beneficial for GGNNs. Accordingly, in this paper, we comprehensively investigate the promotion of GGNNs' capability with the knowledge learned by protein language models (see Fig. 1). The improvements come from two major lines. Firstly, GGNNs can benefit from the information that emerges in the learned representations of those protein language models on fundamental properties of proteins, including secondary structures, contacts, and biological activity. This kind of knowledge may be difficult for GGNNs to be aware of and learn in a specific downstream task. To confirm this claim, we conduct a toy experiment to demonstrate that conventional graph connectivity mechanisms prevent existing GGNNs from being cognizant of residues' absolute and relative positions in the protein sequence. Secondly and more intuitively, protein language models serve as an alternative way of enriching GGNNs' training data and allow GGNNs to be exposed to more different families of proteins, thereby greatly strengthening GGNNs' generalization capability.

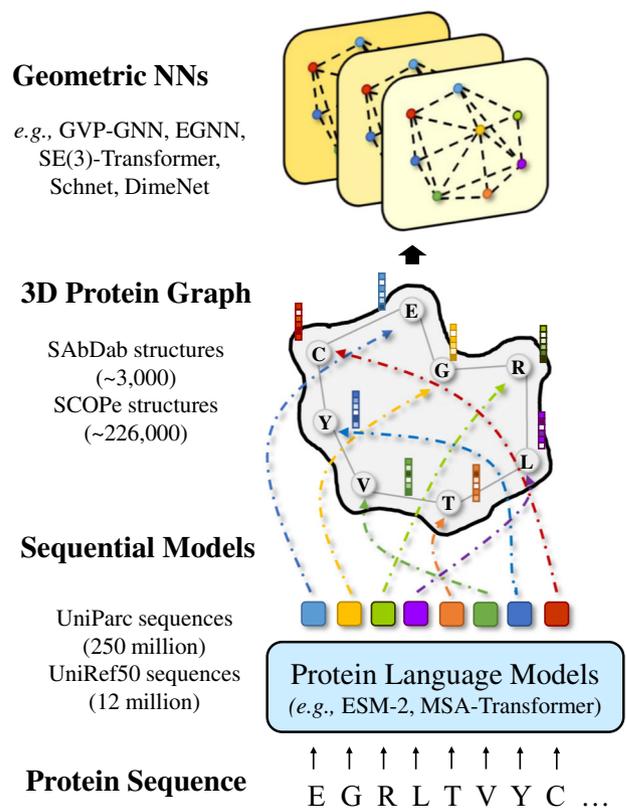

**Fig. 1 Illustration of our framework to strengthen GGNNs with knowledge of protein language models.** The protein sequence is first forwarded into a pretrained protein language model to extract per-residue representations, which are then used as node features in 3D protein graphs for GGNNs.

We examine our hypothesis across a wide range of benchmarks, containing model quality assessment, protein-protein interface prediction, protein-protein rigid-body docking, and ligand binding affinity prediction. Extensive experiments show that the incorporation and combination of pretrained protein language models' knowledge significantly improve GGNNs' performance for various problems, which require distinct domain knowledge. By utilizing the unprecedented view into the language of protein sequences provided by powerful protein language models, GGNNs promise to augment our understanding of a vast database of poorly understood protein structures. Our work hopes to shed more light on how to bridge the gap between the thriving geometric deep learning and mature protein language models and better leverage different modalities of proteins.

## Results and discussion
Our toy experiments illustrate that existing GGNNs are unaware of the positional order inside the protein sequences. Taking a step further, we show in this section that incorporating knowledge learned by large-scale protein language models can robustly enhance GGNN's capacity in a wide variety of downstream tasks.

**Tasks and datasets.**

- **Model Quality Assessment** (MQA) aims to select the best structural model of a protein from a large pool of candidate structures and is an essential step in structure prediction[26]. For a number of recently solved but unreleased structures, structure generation programs produce a large number of candidate structures. MQA approaches are evaluated by their capability of predicting the global distance test (GDT-TS score) of a candidate structure compared to the experimentally solved structure of that target. Its database is composed of all structural models submitted to the Critical Assessment of Structure Prediction (CASP)[27] over the last 18 years. The data is split temporally by competition year. MQA is similar to the Protein Structure Ranking (PSR) task introduced by Townshend et al.[2].
- **Protein-protein Rigid-body Docking** (PPRD) computationally predicts the 3D structure of a protein-protein complex from the individual unbound structures. It assumes that no conformation change within the proteins happens during binding. We leverage Docking Benchmark 5.5 (DB5.5)[28] as the database. It is a gold standard dataset in terms of data quality and contains 253 structures.
- **Protein-protein Interface** (PPI) investigates whether two amino acids will contact when their respective proteins bind. It is an important problem in understanding how proteins interact with each other, e.g., antibody proteins recognize diseases by binding to antigens. We use the Database of Interacting Protein Structures (DIPS), a comprehensive dataset of protein complexes mined from the PDB[29], and randomly select 15K samples for evaluation.
- **Ligand Binding Affinity** (LBA) is an essential task for drug discovery applications. It predicts the strength of a candidate drug molecule's interaction with a target protein. Specifically, we aim to forecast $pK = -\log_{10} K$, where $K$ is the binding affinity in Molar units. We use the PDBbind database[30,31], a curated database containing protein-ligand complexes from the PDB and their corresponding binding strengths. The protein-ligand complexes are split such that no protein in the test dataset has more than 30% or 60% sequence identity with any protein in the training dataset.

**Experimental setup.** We evaluate our proposed framework on the instances of several state-of-the-art geometric networks, using Pytorch[32] and PyG[33] on four standard protein benchmarks. For MQA, PPI, and LBA, we use GVP-GNN, EGNN, and Molformer as backbones. For PPRD, we utilize a deep learning model, EquiDock[34], as the backbone. It approximates the binding pockets and obtains the docking poses using keypoint matching and alignment. For more experimental details, please refer to Supplementary Note 3.

*Single-protein representation task*. For MQA, we document First Rank Loss, Spearman correlation ($R_S$), Pearson's correlation ($R_P$), and Kendall rank correlation ($K_R$) in Table 1. The introduction of protein language models has brought a significant average increase of 32.63% and 55.71% in global and mean $R_S$, of 34.66% and 58.75% in global and mean $R_P$, and of 43.21% and 63.20% in global and mean $K_R$ respectively. With the aid of language models, GVP-GNN achieves the optimal global $R_S$, global $R_P$, and $K_R$ of 84.92%, 85.44%, and 67.98% separately.

Apart from that, we provide a full comparison with all existing approaches in Table 2. We elect RWplus[35], ProQ3D[36], VoroMQA[37], SBROD[38], 3DCNN[2], 3DGNN[2], 3DOCNN[39], DimeNet[40], GraphQA[41], and GBPNet[42] as the baselines. Performance is recorded in Table 2, where the second best is underlined. It can be concluded that even if GVP-GNN is not the best architecture, it can largely outperform existing methods including the state-of-the-art no-pretraining method set by Ayken and Xia[42] (i.e., GBPNet) and the state-of-the-art pretraining results set by Jing et al.[43] if it is enhanced by the protein language model.

*Protein-protein representation tasks*. For PPRD, we report three items as measurements: the complex root mean squared deviation (RMSD), the ligand RMSD, and the interface RMSD in Table 3. The interface is determined with a distance threshold less than 8Å. It is noteworthy that, unlike the EquiDock paper, we do not apply the Kabsch algorithm to superimpose the receptor and the ligand. Contrastingly, the receptor protein is fixed during evaluation. All three metrics decrease considerably with improvements of 11.61%, 12.83%, and 31.01% in complex, ligand, and interface median RMSD, respectively. Notably, we also report the result of EquiDock, which is first pretrained on DIPS and then fine-tuned on DB5. It can be discovered that DIPS-pretrained EquiDock still performs worse than EquiDock equipped with pretrained language models. This strongly demonstrates that structural pretraining for GGNNs may not benefit GGNNs more than pretrained protein language models.

For PPI, we record AUROC as the metric in Fig. 2. It can be found that AUROC increases for 6.93%, 14.01%, and 22.62% for GVP-GNN, EGNN, and Molformer respectively. It is worth noting that Molformer falls behind EGNN and GVP-GNN originally in this task. But after injecting knowledge learned by protein language models, Molformer achieves competitive or even

**Table 1 Results on MQA.**

| Model | PLM | Model quality assessment | | | | | | |
|---|---|---|---|---|---|---|---|---|
| | | First rank loss↓ | Spearman correlation↑ | | Pearson's correlation↑ | | Kendall rank↑ | |
| | | | Mean | Global | Mean | Global | Mean | Global |
| GVP-GNN | ✗ | 0.085 ± 0.002 | 0.4144 ± 0.010 | 0.6910 ± 0.008 | 0.5235 ± 0.013 | 0.6875 ± 0.006 | 0.2960 ± 0.010 | 0.4959 ± 0.004 |
| | ✓ | **0.033 ± 0.001** | **0.6121 ± 0.017** | **0.8492 ± 0.015** | **0.7399 ± 0.017** | **0.8544 ± 0.009** | **0.4530 ± 0.008** | **0.6798 ± 0.014** |
| EGNN | ✗ | 0.054 ± 0.003 | 0.4249 ± 0.016 | 0.7341 ± 0.015 | 0.5315 ± 0.008 | 0.7336 ± 0.018 | 0.3004 ± 0.013 | 0.5344 ± 0.011 |
| | ✓ | **0.041 ± 0.001** | **0.5642 ± 0.013** | **0.8436 ± 0.012** | **0.6925 ± 0.006** | **0.8456 ± 0.015** | **0.4105 ± 0.014** | **0.6558 ± 0.006** |
| Molformer | ✗ | 0.149 ± 0.003 | 0.1238 ± 0.011 | 0.3921 ± 0.004 | 0.1969 ± 0.004 | 0.3901 ± 0.012 | 0.0841 ± 0.010 | 0.2696 ± 0.005 |
| | ✓ | **0.088 ± 0.002** | **0.2424 ± 0.015** | **0.6516 ± 0.009** | **0.3850 ± 0.011** | **0.6210 ± 0.014** | **0.1681 ± 0.012** | **0.4579 ± 0.007** |

The column of 'PLM' indicates whether the protein language model is used. The First Rank Loss is the average difference between the true scores of the best model and the top-ranked model for each target. Results are reported with mean ± standard deviation over three repeated runs and the best performance is in bold.

Table 2 Comparison of performance on MQA.

| Model | PLM | Model quality assessment | | | | | |
|---|---|---|---|---|---|---|---|
| | | Spearman correlation↑ | | Pearson's correlation↑ | | Kendall rank↑ | |
| | | Mean | Global | Mean | Global | Mean | Global |
| RWplus[35a] | ✗ | 0.167 | 0.056 | 0.192 | 0.033 | 0.137 | 0.011 |
| ProQ3D[36a] | ✗ | 0.432 | 0.772 | 0.444 | 0.796 | 0.304 | 594 |
| VoroMQA[37a] | ✗ | 0.419 | 0.651 | 0.412 | 0.651 | 0.291 | 0.505 |
| SBROD[38a] | ✗ | 0.413 | 0.569 | 0.431 | 0.551 | 0.291 | 0.393 |
| 3DOCNN[39b] | ✗ | 0.432 | 0.796 | 0.444 | 0.772 | 0.304 | 0.594 |
| DimeNet[40a] | ✗ | 0.351 | 0.625 | 0.302 | 0.614 | 0.285 | 0.431 |
| 3DCNN[2b] | ✗ | 0.431 ± 0.013 | 0.789 ± 0.017 | 0.557 ± 0.011 | 0.780 ± 0.016 | 0.308 ± 0.010 | 0.592 ± 0.016 |
| 3DGNN[2b] | ✗ | 0.411 ± 0.006 | 0.750 ± 0.018 | 0.500 ± 0.012 | 0.747 ± 0.018 | 0.278 ± 0.005 | 0.547 ± 0.016 |
| GVP-GNN[7c] | ✗ | 0.414 ± 0.010 | 0.691 ± 0.008 | 0.523 ± 0.013 | 0.687 ± 0.006 | 0.296 ± 0.010 | 0.495 ± 0.004 |
| GraphQA[41a] | ✗ | 0.379 | 0.820 | 0.357 | 0.821 | 0.331 | 0.618 |
| GBPNet[42a] | ✗ | <u>0.517</u> | **0.856** | <u>0.612</u> | <u>0.853</u> | <u>0.372</u> | <u>0.656</u> |
| GVP-GNN | ✓ | **0.612 ± 0.017** | <u>0.849 ± 0.015</u> | **0.739 ± 0.017** | **0.854 ± 0.009** | **0.453 ± 0.008** | **0.679 ± 0.014** |

Models are sorted by the year they are released. Results are reported with mean ± standard deviation over three repeated runs and the best and second best performance are bolded and underlined, respectively.
[a]These results are taken from ref. [42].
[b]These results are taken from ref. [2].
[c]These results are re-produced.

Table 3 Performance of PPRD on DB5.5 Test Set.

| Model | PLM | Protein-protein rigid-body docking | | | | | | | | |
|---|---|---|---|---|---|---|---|---|---|---|
| | | Complex RMSD ↓ | | | Ligand RMSD ↓ | | | Interface RMSD ↓ | | |
| | | Median | Mean | Std | Median | Mean | Std | Median | Mean | Std |
| EquiDock | ✗♣ | 16.88 | 17.11 | 5.33 | 40.35 | 37.97 | **12.94** | 16.19 | 37.97 | 4.47 |
|  | ✗♠ | 15.02 | 14.31 | 5.28 | 36.82 | 35.95 | 13.18 | 14.37 | 35.68 | **4.12** |
|  | ✓♣ | **14.92** | **13.14** | **4.59** | **35.17** | **33.48** | 14.34 | **11.17** | **33.48** | 4.38 |

Models with ♣ are directly trained and tested on DB5, while EquiDock with ♠ is first pretrained on DIPS and fine-tuned on the DB5 training set. Results are reported with mean ± standard deviation over three repeated runs and the best performance is in bold.

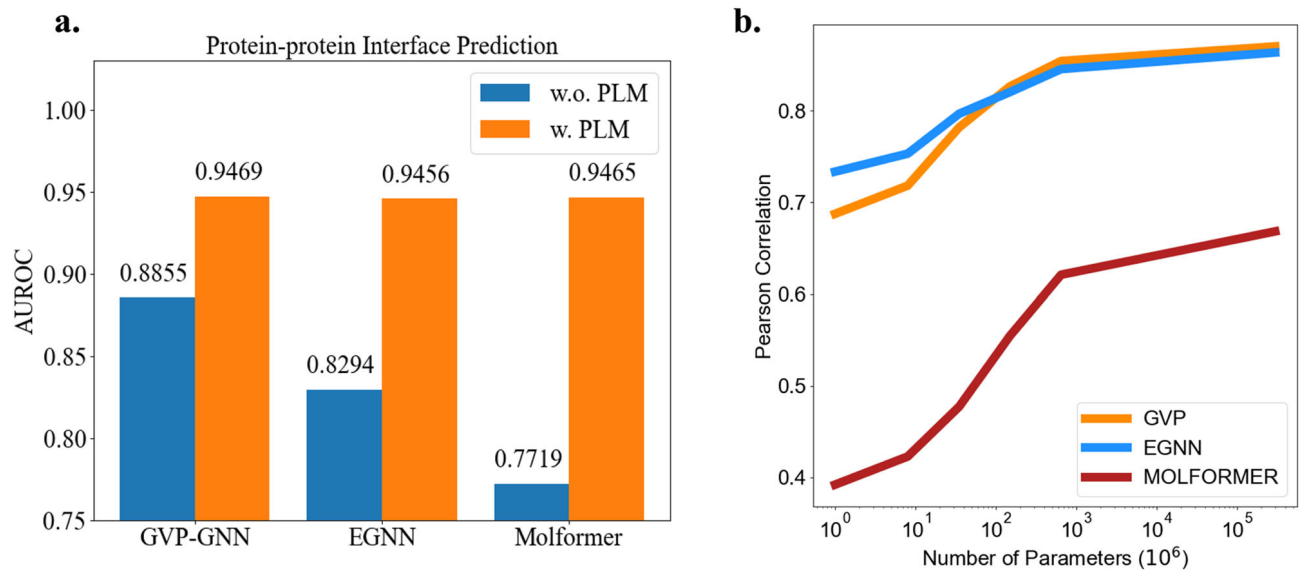

**Fig. 2 Some ablation studies. a** Results of PPI with and without PLMs. **b** Performance of GGNNs on MQA with ESM-2 at different scales.

better performance than EGNN or GVP-GNN. This indicates that protein language models can realize the potential of GGNNs to the full extent and greatly narrow the gap between different geometric deep learning architectures. The results mentioned above are amazing because, unlike MQA, PPRD and PPI study the geometric interactions between two proteins. Though existing protein language models are all trained on single protein sequences, our experiments show that the evolution information hidden in unpaired sequences can also be valuable to analyze the multi-protein environment.

Table 4 Results on LBA.

| Model | PLM | Ligand binding affinity | | | |
|---|---|---|---|---|---|
| | | Sequence identity (30%) | | | |
| | | RMSD↓ | Pearson's correlation↑ | Spearman correlation↑ | Kendall rank↑ |
| GVP-GNN | ✗ | 1.6480 ± 0.014 | 0.2138 ± 0.013 | 0.1648 ± 0.009 | 0.1107 ± 0.012 |
| | ✓ | **1.4556 ± 0.011** | **0.5373 ± 0.010** | **0.5078 ± 0.005** | **0.3495 ± 0.009** |
| EGNN | ✗ | 1.4929 ± 0.012 | 0.4891 ± 0.017 | 0.4725 ± 0.008 | 0.3291 ± 0.014 |
| | ✓ | **1.4033 ± 0.013** | **0.5655 ± 0.016** | **0.5448 ± 0.005** | **0.3790 ± 0.007** |
| Molformer | ✗ | 1.91f07 ± 0.018 | 0.4618 ± 0.014 | 0.4104 ± 0.011 | 0.2812 ± 0.019 |
| | ✓ | **1.6028 ± 0.020** | **0.5351 ± 0.017** | **0.5372 ± 0.015** | **0.3758 ± 0.016** |
| | | Sequence identity (60%) | | | |
| GVP-GNN | ✗ | 1.5438 ± 0.015 | 0.6608 ± 0.012 | 0.6668 ± 0.0010 | 0.4797 ± 0.014 |
| | ✓ | **1.5137 ± 0.019** | **0.6680 ± 0.010** | **0.6716 ± 0.008** | **0.4786 ± 0.012** |
| EGNN | ✗ | 1.5928 ± 0.020 | 0.6274 ± 0.013 | 0.6271 ± 0.017 | 0.4498 ± 0.014 |
| | ✓ | **1.5595 ± 0.022** | **0.6445 ± 0.015** | **0.6463 ± 0.019** | **0.4656 ± 0.019** |
| Molformer | ✗ | 1.8610 ± 0.018 | 0.5528 ± 0.016 | 0.5309 ± 0.015 | 0.3738 ± 0.017 |
| | ✓ | **1.5926 ± 0.024** | **0.6524 ± 0.018** | **0.6528 ± 0.016** | **0.4367 ± 0.011** |

Results are reported with mean ± standard deviation over three repeated runs and the best performance is in bold.

Table 5 Comparison of performance on LBA.

| Model | PLM | Ligand binding affinity (Sequence identity = 30%) | | | |
|---|---|---|---|---|---|
| | | RMSD↓ | Pearson's correlation↑ | Spearman correlation↑ | Kendall rank↑ |
| DeepAffinity[44]a | ✗ | 1.893 ± 0.650 | 0.415 | 0.426 | - |
| Cormorant[45]b | ✗ | 1.568 ± 0.012 | 0.389 | 0.408 | - |
| LSTM[46]c | ✗ | 1.985 ± 0.006 | 0.165 ± 0.006 | 0.152 ± 0.024 | - |
| TAPE[47]c | ✗ | 1.890 ± 0.035 | 0.338 ± 0.044 | 0.286 ± 0.124 | - |
| ProtTrans[14]c | ✗ | 1.544 ± 0.015 | 0.438 ± 0.053 | 0.434 ± 0.058 | - |
| 3DCNN[2]a | ✗ | 1.414 ± 0.021 | 0.550 | 0.553 | - |
| GNN[2]a | ✗ | 1.570 ± 0.025 | 0.545 | 0.533 | - |
| MaSIF[48]c | ✗ | 1.484 ± 0.018 | 0.467 ± 0.020 | 0.455 ± 0.014 | - |
| DGAT[49]b | ✗ | 1.719 ± 0.047 | 0.464 | 0.472 | - |
| DGIN[49]b | ✗ | 1.765 ± 0.076 | 0.426 | 0.432 | - |
| DGAT-GCN[49]b | ✗ | 1.550 ± 0.017 | 0.498 | 0.496 | - |
| GVP-GNN[7]d | ✗ | 1.648 ± 0.014 | 0.213 ± 0.013 | 0.164 ± 0.009 | 0.110 ± 0.012 |
| EGNN[58]d | ✗ | 1.492 ± 0.012 | 0.489 ± 0.017 | 0.472 ± 0.008 | 0.329 ± 0.014 |
| HoloProt[50]c | ✗ | 1.464 ± 0.006 | 0.509 ± 0.002 | 0.500 ± 0.005 | - |
| GBPNet[42]b | ✗ | 1.405 ± 0.009 | 0.561 | **0.557** | - |
| EGNN | ✓ | **1.403 ± 0.013** | **0.565 ± 0.016** | 0.544 ± 0.005 | **0.379 ± 0.007** |

Models are sorted by the year they are released. Results are reported with mean ± standard deviation over three repeated runs and the best and second best performance are bolded and underlined, respectively.
aThese results are taken from ref. 2.
bThese results are taken from ref. 42.
cThese results are copied from ref. 50.
dThese results are re-produced.

*Protein-molecules representation task*. For LBA, we compare RMSD, $R_S$, $R_P$, and $K_R$ in Table 4. The incorporation of protein language models produces a remarkably average decline of 11.26% and 6.15% in RMSD for 30% and 60% identity, an average increase of 51.09% and 9.52% in $R_P$ for the 30% and 60% identity, an average increment of 66.60% and 8.90% in $R_S$ for the 30% and 60% identity, and an average increment of 68.52% and 6.70% in $K_R$ for the 30% and 60% identity. It can be seen that the improvements in the 30% sequence identity is higher than that in the less restrictive 60% sequence identity. This confirms that protein language models benefit GGNNs more when the unseen samples belong to different protein domains. Moreover, contrasting PPRD or PPI, LBA studies how proteins interact with small molecules. Our outcome demonstrates that rich protein representations encoded by protein language models can also contribute to the analysis of protein's reaction to other non-protein drug-like molecules. The result of a different data split has been placed in Supplementary Table 1.

In addition, we compare thoroughly with existing approaches for LBA in Table 5, where the second best is underlined. We select a broad range of models including DeepAffinity[44], Cormorant[45], LSTM[46], TAPE[47], ProtTrans[14], 3DCNN[2], GNN[2], MaSIF[48], DGAT[49], DGIN[49], DGAT-GCN[49], HoloProt[50], and GBPNet[42] as the baseline. It is clear that even if EGNN is a median-level architecture, it can achieve the best RMSD and the best Pearson's correlation when enhanced by protein language models, beating a group of strong baselines including HoloProt[50] and GBPNet[42].

**Scale and type of protein language models**. It has been observed that as the size of the language model increases, there are consistent improvements in tasks like structure prediction[12]. Here we conduct an ablation study to investigate the effect of protein language models' sizes on GGNNs. Specifically, we explore different ESM-2 with the parameter numbers of 8M, 35M, 150M, 650M, and 3B and plot results in Fig. 2. It verifies that scaling the protein language model is advantageous for GGNNs. More additional results can be found in Supplementary Note 4. We also provide a comparison of different sorts of PLMs' influence in Supplementary Table 2. Besides that, we investigate the difference of PLMs' effectiveness with and without MSA in Supplementary Table 3.

**Limitations.** Despite our successful confirmation that PLMs can promote geometric deep learning, there are several limitations and extensions of our framework left open for future investigation. For instance, our 3D protein graphs are residue-level. We believe atom-level protein graphs also benefit from our approach, but its increase in performance needs further exploration.

## Conclusion
In this study, we investigate a problem that has been long ignored by existing geometric deep learning methods for proteins. That is, how to employ the abundant protein sequence data for 3D geometric representation learning. To answer this question, we propose to leverage the knowledge learned by existing advanced pre-trained protein language models and use their amino acid representations as the initial features. We conduct a variety of experiments such as protein-protein docking and model quality assessment to demonstrate the efficacy of our approach. Our work provides a simple but effective mechanism to bridge the gap between 1D sequential models and 3D geometric neural networks, and hope to throw light on how to combine information encoded in different protein modalities.

## Method
### Sequence recovery analysis
*Preliminary and motivations.* It is commonly acknowledged that protein structures maintain much more information than their corresponding amino acid sequences. And for decades long, it has been an open challenge for computational biologists to predict protein structure from its amino acid sequence. Though the advancement of Alphafold (AF)[51] and RosettaFold[52] has made a huge step in alleviating the limitation brought by the number of available experimentally determined protein structures, neither AF nor its successors such as Alphafold-Multimer[53], IgFold[54], and HelixFold[55] are a panacea. Their predicted structures can be severely inaccurate when the protein is orphan and lacks multiple sequence alignment (MSA) as the template. Consequently, it is hard to conclude that protein sequences can be perfectly transformed to the structure modality by current tools and be used as extra training resources for GGNNs.

Moreover, we argue that even if conformation is a higher-dimensional representation, the prevailing learning paradigm may forbid GGNNs from capturing the knowledge that is uniquely preserved in protein sequences. Recall that GGNNs are mainly diverse in their patterns to employ 3D geometries, the input features include distance[56], angles[40], torsion, and terms of other orders[57]. The position index hidden in protein sequences, however, is usually neglected when constructing 3D graphs for GGNNs. Therefore, in this section, we design a toy trial to examine whether GGNNs can succeed in recovering this kind of positional information.

*Protein graph construction.* Here the structure of a protein can be represented as an atom-level or residue-level graph $\mathcal{G} = (\mathcal{V}, \mathcal{E})$, where $\mathcal{V}$ and $\mathcal{E} = (e_{ij})$ correspond to the set of $N$ nodes and $M$ edges respectively. Nodes have their 3D coordinates $\mathbf{x} \in \mathbb{R}^{N \times 3}$ and the initial $\psi_h$-dimension roto-translational invariant features $\mathbf{h} \in \mathbb{R}^{N \times \psi_h}$ (e.g., atom types and electronegativity, residue classes). Normally, there are three types of options to construct connectivity for molecules: *r-ball* graphs, *fully-connected* (FC) graphs, and *K-nearest neighbors* (KNN) graphs. In our setting, nodes are linked to $K = 10$ nearest neighbors for KNN graphs, and edges include all atom pairs within a distance cutoff of 8Å for r-ball graphs.

*Recovery from graphs to sequences.* Since most prior studies choose to establish 3D protein graphs based on purely geometric information and ignore their sequential identities, it provokes the following position identity question:

*Can existing GGNNs identify the sequential position order only from geometric structures of proteins?*

To answer this question, we formulate two categories of toy tasks (see Fig. 3). The first one is absolute position recognition (APR), which is a classification task. Models are asked to directly predict the position index ranging from 1 to $N$, the residue number of each protein. This task adopts accuracy as the metric and expects models to discriminate the absolute position of the amino acid within the whole protein sequence. We compute the distribution of the protein sequence lengths in Supplementary Fig. 1.

In addition to that, we propose the second task named relative position estimation (RPE) to focus on the relative position of each residue. Models are required to predict the minimum distance of residue to the two sides of the given protein and the root mean squared error (RMSE) is used as the metric. This task aims to examine the capability of GGNNs to distinguish which segment the amino acid belongs to (i.e., the center section of the protein or the end of the protein).

*Experiments*

Backbones: We adopt three technically distinct and broadly accepted architectures of GGNNs for empirical verification. To be specific, GVP-GNN[7,43] extends standard dense layers to operate on collections of Euclidean vectors, performing both geometric and relational reasoning on efficient representations of macromolecules. EGNN[58] is a translation, rotation, reflection, and permutation equivariant GNN without expensive spherical harmonics. Molformer[9] employs the self-attention mechanism for 3D point clouds while guarantees SE(3)-equivariance.

Dataset: We exploit a small non-redundant subset of high-resolution structures from the PDB. To be specific, we use only X-ray structures with resolution < 3.0Å, and enforce a 60% sequence identity threshold. This results in a total of 2643, 330, and 330 PDB structures for the train, validation, and test sets, respectively. Experimental details, the summary of the database, and the description of these GGNNs are elaborated in Supplementary Notes 1 and 2.

Empirical results and analysis: Table 6 documents the overall results, where metrics are labeled with ↑/↓ if higher/lower is better, respectively. It can be found that all GGNNs fail to recognize either the absolute or the relative positional information encoded in the protein sequences with an accuracy lower than 1% and an extremely high RMSE.

This phenomenon stems from the conventional ways to build graph connectivity, which usually excludes sequential information. To be specific, unlike common applications of GNNs such as citation networks[59], social networks[60], knowledge graphs[61], molecules do not have explicitly defined edges or adjacency. On the one hand, r-ball graphs utilize a cut-off distance, which is usually set as a hyperparameter, to determine the particle connections. But it is hard to guarantee a cut-off to properly include all crucial node interactions for complicated and large molecules. On the other hand, FC graphs that consider all pairwise distances will cause severe redundancies, dramatically increasing the computational complexity especially when proteins consist of thousands of residues. Besides, GGNNs also

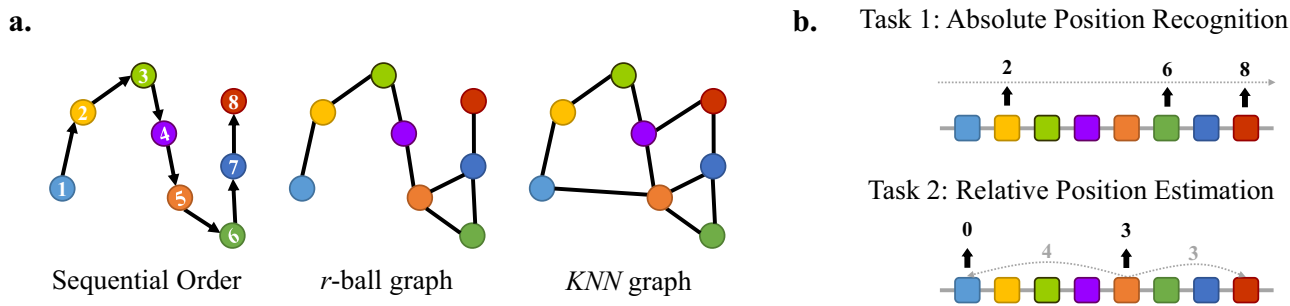

**Fig. 3 Illustration of the sequence recovery problem. a** Protein residue graph construction. Here we draw graphs in 2D for better visualization but study 3D graphs for GGNNs. **b** Two sequence recovery tasks. The first requires GGNNs to predict the absolute position index for each residue in the protein sequence. The second aims to forecast the minimum distance of each amino acid to the two sides of the protein sequence.

**Table 6 Results of two residue position identification tasks.**

| Models | Graph type | APR | RPE |
|---|---|---|---|
| | | Accuracy (%) ↑ | RMSE ↓ |
| GVP-GNN | r-ball graph | 0.157 ± 0.002 | 392.38 ± 3.41 |
| | KNN graph | **0.158 ± 0.003** | 392.38 ± 4.05 |
| EGNN | r-ball graph | 0.150 ± 0.005 | 412.70 ± 2.36 |
| | KNN graph | 0.131 ± 0.004 | 403.86 ± 1.77 |
| Molformer | FC graph | 0.148 ± 0.007 | **270.69 ± 4.53** |

Results are reported with mean ± standard deviation over three repeated runs and the best performance is in bold.

easily get confused by excessive noise, leading to unsatisfactory performance. As a remedy, KNN becomes a more popular choice to establish graph connectivity for proteins[34,62,63]. However, all of them take no account of the sequential information and require GGNNs to learn this original sequential order during training.

The lack of sequential information can yield several problems. To begin with, residues are unaware of their relative positions in the proteins. For instance, two residues can be close in the 3D space but distant in the sequence, which can mislead models to find the correct backbone chain. Secondly, according to the characteristics of the MP mechanism, two residues in a protein with the same neighborhood are expected to share similar representations. Nevertheless, the role of those two residues can be significantly separate[64] when they are located at different segments of the protein. Thus, GGNNs may be incapable of differentiating two residues with the same 1-hop local features. This restriction has already been distinguished by several works[6,65], but none of them make a strict and thorough investigation. Admittedly, sequential order may only be necessary for certain tasks. But this toy experiment strongly indicates that the knowledge monopolized by amino acid sequences can be lost if GGNNs only learn from protein structures.

**Integration of language models into geometric networks.** As discussed before, learning about 3D structures cannot directly benefit from large amounts of sequential data. Subsequently, the model sizes of GGNNs are limited, or instead, overfitting may occur[66]. On the contrary, comparing the number of protein sequences in the UniProt database[67] to the number of known structures in the PDB, there are over 1700 times more sequences than structures. More importantly, the availability of new protein sequence data continues to far outpace the availability of experimental protein structure data, only increasing the need for accurate protein modeling tools.

Therefore, we introduce a straightforward approach to assist GGNNs with pretrained protein language models. To this end, we feed amino acid sequences into those protein language models, where ESM-2[12] is adopted in our case, and extract the per-residue representations, denoted as $h' \in \mathbb{R}^{N \times \psi_{PLM}}$. Here $\psi_{PLM} = 1280$. Then $h'$ can be added or concatenated to the per-atom feature $h$. For residue-level graphs, $h'$ immediately replaces the original $h$ as the input node features.

Notably, incompatibility exists between the experimental structure and its original amino acid sequence. That is, structures stored in the PDB files are usually incomplete and some strings of residues are missing due to inevitable realistic issues[68]. They, therefore, do not perfectly match the corresponding sequences (i.e., FASTA sequence). There are two choices to address this mismatch. On the one hand, we can simply use the fragmentary sequence as the substitute for the integral amino acid sequence and forward it into the protein language models. On the other hand, we can leverage a dynamic programming algorithm provided by Biopython[69] to implement pairwise sequence alignment and abandon residues that do not exist in the PDB structures. It is empirically discovered that no big difference exists between them, so we adopt the former processing mechanism for simplicity.

**Reporting summary**. Further information on research design is available in the Nature Portfolio Reporting Summary linked to this article.

## Data availability

The data of model quality assessment, protein-protein interface prediction, and ligand affinity prediction is available by https://www.atom3d.ai/. The data of protein-protein rigid-body docking can be downloaded directly from the official repository of Equidock https://github.com/octavian-ganea/equidock_public. Source data for figures can be found in Supplementary Data.

## Code availability

The code repository is stored at https://github.com/smiles724/bottleneck. It is also deposited in ref.[70].

## Acknowledgements
This work is supported in part by the Institute of AI Industry Research at Tsinghua University and the Molecule Mind.

## Author contributions
F.W. and J.X. led the research. F.W. contributed technical ideas. F.W. and Y.T. developed the proposed method. F.W., D.R., and Y.T. performed the analysis. J.X. and D.R. provided evaluation and suggestions. All authors contributed to the manuscript.

## Competing interests
The authors declare no competing interests.

## Additional information
**Supplementary information** The online version contains supplementary material available at https://doi.org/10.1038/s42003-023-05133-1.

**Correspondence** and requests for materials should be addressed to Stan Z. Li.

**Peer review information** *Communications Biology* thanks Jianzhao Gao, Arne Elofsson, and the other, anonymous, reviewer(s) for their contribution to the peer review of this work. Primary Handling Editors: Yuedong Yang and Gene Chong.

**Reprints and permission information** is available at http://www.nature.com/reprints

**Publisher's note** Springer Nature remains neutral with regard to jurisdictional claims in published maps and institutional affiliations.